# Generating Progressive Images from Pathological Transitions via Diffusion Model

Zeyu Liu[1,a], Tianyi Zhang[1,a], Yufang He[a], Yunlu Feng[b], Yu Zhao[c], Guanglei Zhang[*,a]

[a]Beijing Advanced Innovation Center for Biomedical Engineering, School of Biological Science and Medical Engineering, Beihang University, Beijing, 100191, China
[b]Department of Gastroenterology, Peking Union Medical College Hospital, Beijing 100006, China
[c]Department of Pathology, Peking Union Medical College Hospital, Beijing 100006, China
[1]Authors contributed equally
[*]Corresponding author: Guanglei Zhang (guangleizhang@buaa.edu.cn)

**Abstract.** Deep learning is widely applied in computer-aided pathological diagnosis, which alleviates the pathologist workload and provide timely clinical analysis. However, most models generally require large-scale annotated data for training, which faces challenges due to the sampling and annotation scarcity in pathological images. The rapid developing generative models shows potential to generate more training samples from recent studies. However, they also struggle in generalization diversity with limited training data, incapable of generating effective samples. Inspired by the pathological transitions between different stages, we propose an adaptive depth-controlled diffusion (ADD) network to generate pathological progressive images for effective data augmentation. This novel approach roots in domain migration, where a hybrid attention strategy guides the bidirectional diffusion, blending local and global attention priorities. With feature measuring, the adaptive depth-controlled strategy ensures the migration and maintains locational similarity in simulating the pathological feature transition. Based on tiny training set (samples $\leq$ 500), the ADD yields cross-domain progressive images with corresponding soft-labels. Experiments on two datasets suggest significant improvements in generation diversity, and the effectiveness with generated progressive samples are highlighted in downstream classifications. The code is available at https://github.com/Rowerliu/ADD.

**Keywords:** Pathological image analysis, Data augmentation, Diffusion models, Image generation

## 1 Introduction

Histopathology widely serves as the gold standard for cancer diagnosis [1], which relies on pathologists to capture the tissue and cellular features of microscopic images [2, 3]. With advances in deep learning, neural networks have been applied to assist pathological diagnosis, and proven to be effective [4, 5]. However, training effective models generally requires a large number of images and high-quality annotations, which severely challenges this field. Moreover, unlike natural images, pathological images generally exhibit higher information densities and similarities [6, 7]. Accordingly, scaling up similar samples contributes limited effectiveness, while learning samples with complex and ambiguous patterns help to capture the prominent features. These characteristics result in the demand for effective training samples that vary across different pathological states, further increasing the difficulty and cost in sample collection and annotation [8].

Nevertheless, recent studies shed insight on synthesizing effective training samples [9,10]. In this generative task, traditional GAN-based models are commonly employed for their interpretable optimization advantages. However, drawbacks in generative diversity and training difficulty hinder their potential in generating high-quality samples [11]. In recent studies [12], the diffusion model serves as a promising alternative. Via likelihood-based generation, diffusion models present desirable properties in distribution coverage, stationary training and easy scalability [13,14]. Still, the lack of effective controlling and the easy overfitting specialty limits diffusion models on large dataset training, which stands against the original purpose of overcoming data deficiencies [15]. Consequently, current methods lay significant limits in balancing training scales and generation diversity, which can be summarized as "generative contradiction".

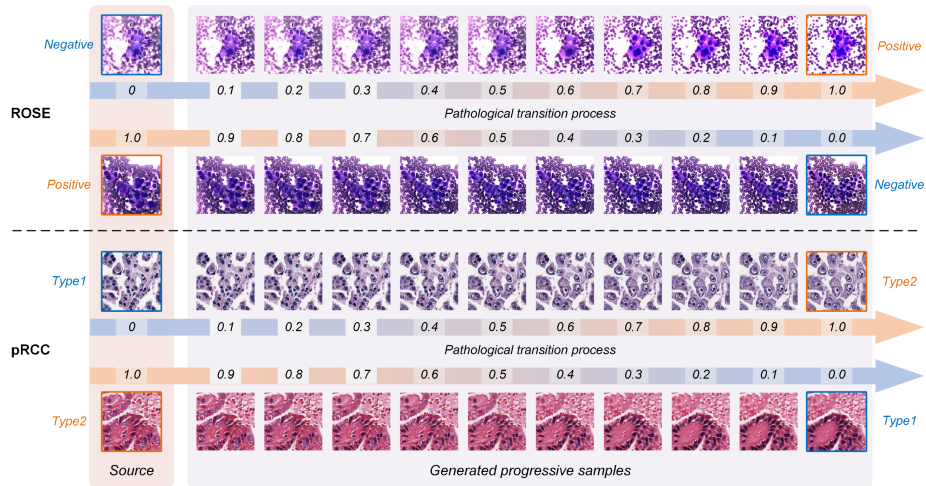

**Fig. 1.** The generated progressive samples with their soft-labels. Progressing between different pathological stages, feature transition is simulated while the locational similarity is maintained.

To resolve this "generative contradiction", we specially focused on progressive characteristics in the pathological stage transition. Specifically, samples of different physiological states reveal feature transition and locational similarity. Therefore, simulating these transitions while maintaining locational similarity yields a series of progressive samples between different pathological states. As shown in **Fig. 1** with two diseases, the progressive samples present between-stage patterns (their classification criteria is illustrated in the Supplementary Materials Fig. 1). These complex progressive patterns make up effective training samples therefore motivates our proposal on adaptive depth-controlled diffusion (ADD) network.

In this work, firstly, a pathological domain migration approach is designed where the diffusion models are trained with two domains. It provides the basis for the connection between different pathological states. Secondly, a novel hybrid attention strategy (HAS) is designed to maintain the effective local and global similarities in generation. This design enables training with limited samples while generating diverse effective samples following prior knowledge of the source domain. Lastly, in addition to the final-state domain migration, we design feature measuring for depth control, which not only ensures image quality regarding the locational similarity, but also allows unlimited progressive cross-domain image generation. With these designs, ADD is able to generate the progressive samples between different pathological states.

In summary, we are one of the first to focus on progressive characteristics in pathological images, where a novel data generation approach ADD is proposed. The hybrid attention and adaptive depth-controlled strategies effectively ensure the migration and generate progressive samples between different pathological stages. Numerical evaluation and case studies prove the advanced generation quality and diversity compared to the state-of-the-art (SOTA) methods. Furthermore, adapt to small datasets, we generate effective samples for progressive soft-labels supervisions. Experimental results on two pathological datasets highlight generation performance and significant enhancement in the downstream classification.

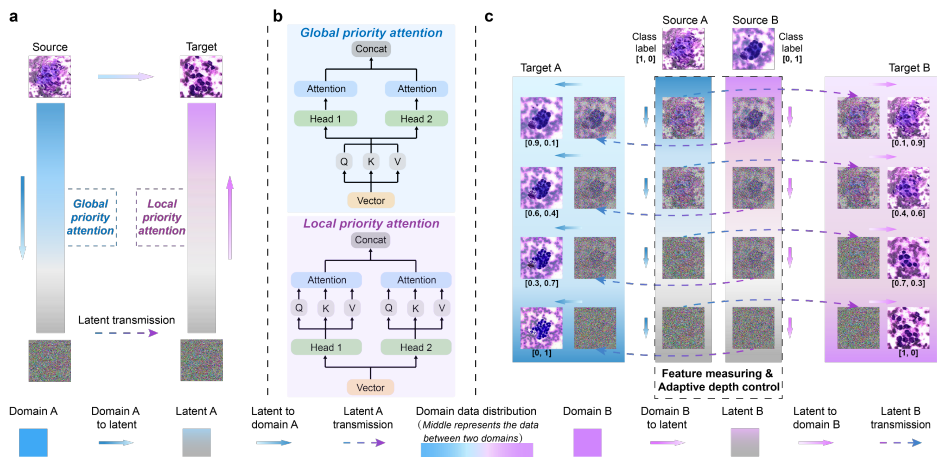

**Fig. 2.** Overview of ADD method, where the U-BDP for domain migration is shown in **a**; a hybrid attention strategy with different local and global focuses is shown in **b**; and an adaptive depth-controlled strategy with feature measuring is shown in **c**.

## 2 Method

### 2.1 Diffusion Models for Domain Migration

To achieve cross-domain migration, we represent a U-shaped bidirectional diffusion process (U-BDP) shown in **Fig. 2a.** Based on the denoising diffusion implicit model (DDIM), it encompasses two basic processes: the forward diffusion process and the reverse denoising process [16,17]. In the forward process, the noisy image $x_t$ and the original image $x_0$ can be expressed as:

$$q(x_t|x_0) = \mathrm{N}\left(\sqrt{\alpha_t}x_0, (1-\alpha_t)\mathbf{1}\right), \tag{1}$$

where $\alpha_t$ is the parameter, and $\mathbf{1}$ is the identity matrix. In the reverse denoising process, $x_{t-1}$ can be obtained by reversing the forward diffusion process to sample from the noisy input $x_0$ with a normal distribution:

$$x_{t-1} = \sqrt{\alpha_{t-1}}\frac{x_t - \sqrt{1-\alpha_t}\varepsilon_\theta(x_t)}{\sqrt{\alpha_t}} + \sqrt{1-\alpha_{t-1}-\sigma_t^2}\varepsilon_\theta(x_t) + \sigma_t\varepsilon. \tag{2}$$

where $\varepsilon_\theta(x_t)$ represents the normally distributed noise predicted by $\theta$ based on the noisy image at step $t$.

This denoising process can be totally definite, if $\sigma_t$ is set to 0, making $x_{t-1}$ independent with random noise $\varepsilon$. This provides a theoretical basis for bridging two different domains [18]. To achieve conciseness, we use the probability flow ordinary differential equation (ODE) for representation. Mapping $x(t_1)$ from $x(t_0)$ can be defined as:

$$ODE\left(x(t_0); v_\theta, t_0, t_1\right) = x(t_0) + \int_{t_0}^{t_1} v_\theta(t, x(t)) dt, \tag{3}$$

where $v_\theta = dx/dt$. Furthermore, image transfer can be achieved by bridging the two domains through a connection of two bidirectional ODE equations:

$$x_A^{(l)} = ODE\left(x_A^{(s)}; v_\theta^{(s)}, 0, 1\right), x_B^{(t)} = ODE\left(x_A^{(l)}; v_\theta^{(t)}, 1, 0\right), \tag{4}$$

Therefore, the latent space representation $x_A^{(l)}$ of the source image $x_A^{(s)}$ is obtained through the forward ODE in domain A, and then transformed into $x_B^{(t)}$ in domain B using the reversed ODE denoising process.

## 2.2 Hybrid Attention Strategy

To ensure the effectiveness of generated sample, the domain migration aims to establish a visual connection between the migrated samples and their source counterparts. Accordingly, the ideal output image $x_B^{(t)}$ should strictly belong to target domain B while maintaining the visual consistency with $x_A^{(s)}$. Based on the U-Net with self-attention [16] for predicting noise in the diffusion training, we further proposed a hybrid attention strategy (HAS) to achieve this intension. As shown in **Fig. 2b**, the self-attention calculation includes two essential steps: obtaining query (Q), key (K), and value (V) and segmenting tokens via multi-head. Accordingly, prioritizing attention calculation through Q, K, V highlight establishing global connections. On the contrary, prioritizing token segmentation via multi-head enhances the local perception capabilities.

In our design, the local and global features are focused in different migration processes. Specifically, we prioritize the global self-attention of the forward process to obtain $x_A^{(l)}$, which retains the appearance consistency. Conversely, in the reverse process, we prioritize local self-attention to capture spatial features in domain B, ensuring the consistency in domain migration with prominent spatial features.

### 2.3 Adaptive Depth-controlled Strategy with Feature Measuring

To effectively ensure the migration and generate progressive samples between different pathological stages, we further propose the adaptive depth-controlled strategy. Due to the consecutiveness of ODE, for any step $t, t \in (0, T)$ corresponding to an intermediate state $i, i = t/T$, the following process still exists:

$$x^{(i)} = ODE\left(x^{(i)}; v_\theta^{(i)}, 0, i\right), x^{(i)} = ODE\left(x^{(i)}; v_\theta^{(i)}, 1, i\right),$$ (5)

Since different ODEs are trained using the same diffusion process, the linear adding noise builds $x_A^{(i)}$ and $x_B^{(i)}$ to have a similar noise level at any state $i$. In domain migration, $x^{(i)}$ serves as an intermediate state between A and B, which is positively correlated to $i/T$. Therefore, it allows us to simulate the pathological transition process when generate the images. As shown in **Fig. 2c**, by controlling the depth of the U-BDP, progressive state of generation can be controlled via the domain migration state $i$. Moreover, the soft-labels can be assigned to describe the states of generated images:

$$L_A, L_B, L_I, \left(L_A = 0, L_B = 1, L_I \in (0,1)\right).$$ (6)

where $L_A$, $L_B$ and $L_I$ represents the label of samples in domain A, domain B and intermediate states.

Generally, depths with substantial distances should yield very similar migration images. However, for a same progressive state $L_I$, the migration state depths $i$ may vary significantly in different samples. Accordingly, we design a feature measuring indicator to control the state depths $i$ and ensure the generation acurracy. Specifically, the pattern of pathological images is primarily determined by the tissue or cellular structures, which are mainly carried by high-frequency details (shown in Supplementary Materials Fig.2). Therefore, the noising and denoising process of diffusion can be viewed as encompassing the degradation and recovery of these high-frequency information.

To measure the feature change, we perform a fast Fourier transform (FFT) on $x^{(i)}$ and then use a high-pass filter through mask $M_h$. Its average magnitude can be then calculated as $A_{x^{(i)}}$:

$$F\left(x^{(i)}\right)_h = FFT\left(x^{(i)}\right) \odot M_h$$ (7)

$$A_{x^{(i)}} = \overline{F\left(x^{(i)}\right)_h}$$ (8)

Given the amplitude of $x^{(i)}$, $x^{(i)}$ and $x^{(i)}$ through the same above way, progressive state $L_I$ corresponding to $x^{(i)}$ can be determined as follows:

$$L_I = \left(A_{x^{(i)}} - A_{x^{(i)}}\right) / \left(A_{x^{(i)}} - A_{x^{(i)}}\right)$$ (9)

Similarly, in the given case of any $x^{(i)}$ and corresponding $x^{(i)}$, it is also possible to generate intermediate pathological image $I$ with the given soft label $L_I$.

## 3 Experiment

### 3.1 Dataset and Experimental Settings

**Dataset.** The pancreatic cancer Rapid On-site Evaluation (ROSE) dataset was collected at the Peking Union Medical College Hospital. It composes of 1154 typical pancreatic cancer samples and 795 normal samples labelled by senior pathologists. The Papillary Renal Cell Carcinoma (pRCC) dataset is a binary-classification (870 Type1 and 547 Type2 samples) dataset open-sourced from study [19]. In this study, 500 samples from each category are randomly selected as training set, while the remaining samples are splitted into validation and test set under a ratio of 1:2. Accordingly, the training set is used in the generation and downstream training, while the validation and test sets are only applied in the downstream evaluations. Full dataset is used for numerical quality evaluation for the generated images.

**Baselines.** In sample generation, we have implemented several SOTA methods following their official settings, including ProGAN, LoFGAN and MixDL based on GAN, and IDDPM based on Diffusion [20-23]. In the downstream classification, ViT is employed as the backbone for downstream classification test. The hyper-parameters of all methods are optimized in experiments.

**Metrics.** To compare the generation performance numerically, following [15,20-23], we employ the Frechet Inception Distance (FID) and sFID to measure the distribution difference of generated and real images. Furthermore, we use the Learned Perceptual Image Patch Similarity (LPIPS) to measure the diversity of generated samples. Lastly, in downstream classification, standard evaluation of accuracy and F1-score are reported.

### 3.2 Generation Evaluation

**Numerical Evaluation:** To evaluate the effectiveness on sample generation, the ADD and SOTAs are utilized to randomly generate 500 pseudo samples in each category. The numerical comparisons are reported in **Table 1**, where the FID and sFID indicate better performance in lower values while LPIPS indicates better performance in higher values.

On both datasets, ADD achieves the best numerical results compared to SOTAs. Specifically, ADD introduces the pathological domain migration approach to model changes in physiological states. Meanwhile, with hybrid attention strategy, pseudo images with high local similarity are generated, resulting in better FID and sFID performance. Moreover, regarding generation diversity, the diffusion-based ADD shows more strength in learning the diverse distribution compared to the traditional GAN-based models. By exploiting the few-shot training samples, ADD provides higher diversity (LPIPS) against other methods.

**Table 1.** Generation Performance Evaluation on ROSE and PRCC datasets.

| Model | ROSE | | | pRCC | | |
|---|---|---|---|---|---|---|
| | FID(↓) | sFID(↓) | LPIPS(↑) | FID(↓) | sFID(↓) | LPIPS(↑) |
| ProGAN[20] | 147.83 | 127.31 | 0.6014 | 129.58 | 133.54 | 0.5892 |
| IDDPM [21] | 74.52 | 83.96 | 0.5109 | 64.52 | 83.96 | 0.5328 |
| LoFGAN [22] | 91.77 | 87.74 | 0.5621 | 83.15 | 95.72 | 0.5534 |
| MixDL [23] | 122.92 | 100.73 | 0.5398 | 117.75 | 88.85 | 0.5013 |
| ADD (no-HAS) | **61.71** | 74.64 | 0.4398 | 67.93 | 84.64 | 0.4125 |
| **ADD** | 67.83 | **70.29** | **0.6267** | **58.31** | **81.26** | **0.6078** |

**Case Study:** ADD and three comparison methods with better numerical performance are presented in **Fig. 3a** Specifically, 4 original images from 2 datasets are shown in the first row, and the generated images are shown in the following rows. With the analysis of senior pathologists, the pseudo samples generated by ADD are visually closer to the original samples, which retain the original cell distribution outlines. Moreover, they present the diverse pathological spatial changes regarding the pathological states. Specifically, compared to other methods, the unique anisocytosis and irregular orientation can be more accurately expressed from ADD in positive ROSE samples. Meanwhile, ADD generates pRCC samples with clear basal cell proliferation and cell layer differences, demonstrating its advanced multi-scale feature modelling. More generated samples are shown in Supplementary Materials Fig. 3-6.

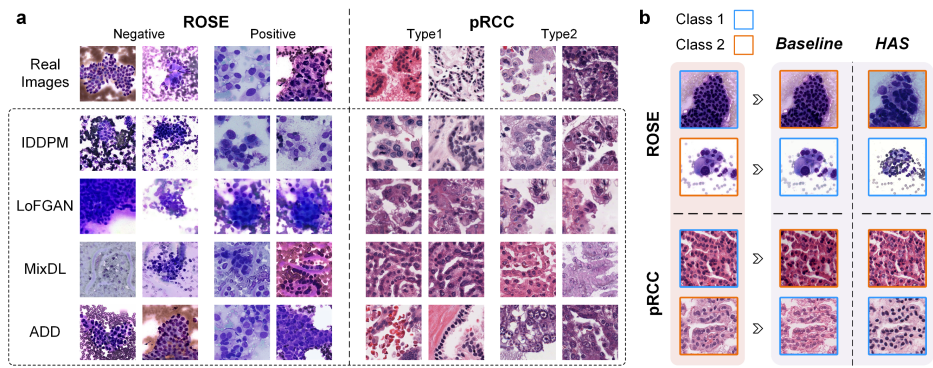

**Fig. 3.** The ablation results on pathological domain migration

**Effective migration with HAS:** The U-BDP provides with the possibility to migrate images from different domains. Nevertheless, pathological images share the similar biological spatial features and progressive global feature distributions, highlighting learning the spatial and global similarities and progressive patterns. Illustrated in **Fig. 3b**, using U-BDP baseline in migration results only minor changes in color and contours. Conversely, the proposed HAS effectively enhances the reverse denoising process with feature recognition between the hidden space and the two source domains. Accordingly, the generated images retain the original color style and spatial distribution, while maintaining consistency with the pattern distribution of the target domain. The numerical performance in **Table 2** confirms the effectiveness of HAS (ADD vs U-BDP), while the generated samples in **Fig. 3b** visualize its performance.

### 3.3 Effectiveness in Generating Progressive Samples

**Generating Progressive Pathological Samples:** To simulate the pathological transition process, an adaptive depth-controlled strategy is designed in ADD. It controls the extent of domain migration by measuring the feature similarity to the target domain. Specifically, feature similarity is measured by the adaptive image Fourier transform, which yield the progressive depth and corresponding classification soft labels. Two series of yielding progressive images are shown in **Fig. 1.** These high-quality progressive pathological samples provide a novel learning objective for downstream tasks.

**Classification Evaluation:** Generating samples is serving as effective data-augmentations for the downstream tasks. Accordingly, we employ all methods to generate 500 new pseudo samples, forming a series generated training sets. Then, these generated training sets are used for comparison under the same conditions with the baseline real-image training set (Supplementary Materials Table 1). Generating the final stage of pseudo samples, ViT trained with ADD outperforms other SOTA generations on both ROSE and pRCC datasets. The results prove that our proposed method generates more effective samples with highest distribution consistency to the real images.

Then, the data augmentation performance is explored in the mixed scenario, where a series of aug-trainsets are assembled with 500 generated samples and 500 real samples. The corresponding classification results are shown in Supplementary Materials Table 2, and some improvements can be observed from IDDPM, LoFGAN and ADD. This validates the augmentation of generative methods for downstream tasks, and shows the superiority of ADD from comparison. We attribute this to the amplified diversity and weakening of model overfitting from data expansion.

However, generation enhancements are limited or even leads to a decrease when scaling up the aug-trainset to 5000 generated samples. This indicates the excessive data may be too similar and redundant for downstream learning. On the contrary, taking the pathological observation on progressive states, ADD is used to generate 500 progressive samples at multiple progressive states, forming a total of 5000 soft-labeled samples as training set. Marked as ADD-Pro in **Table 2**, it demonstrates far superior data augmentation performance, evidently enhanced by the most effective samples via progressive data generation.

Table 2. Comparisons in the downstream classifications using aug-trainset (5000). The arrow indicates the data augmentation of generated models.

| Generative | ROSE | | pRCC | |
|---|---|---|---|---|
| Model | Accuracy (%) | F1-score (%) | Accuracy (%) | F1-score (%) |
| Baseline | 78.13 | 79.09 | 83.50 | 85.36 |
| ProGAN | 71.01 | 71.58 | 73.69 | 74.01 |
| IDDPM | 78.21↑ | 79.25↑ | 84.10↑ | 85.42↑ |
| LoFGAN | 73.98 | 74.02 | 83.65↑ | 85.48↑ |
| MixDL | 66.35 | 68.25 | 76.28 | 77.26 |
| ADD (Ours) | 79.87↑ | 80.53↑ | 86.02↑ | 87.54↑ |
| ADD-Pro (Ours) | **81.01**↑ | **83.14**↑ | **87.45**↑ | **88.77**↑ |

## 4 Conclusion

In conclusion, with the observation of the pathological transitions, a novel adaptive depth-controlled diffusion (ADD) network is proposed to generate progressive samples. With the hybrid attention strategy and adaptive depth-controlled strategy, effective pathological images are generated through domain migration, overcoming the "generative contradiction" with high generation quality and diversity. Moreover, by applying generated progressive samples in downstream classification, we are one of the first to explore the progressive pathological states as an effective learning objective. Numerical SOTA performance brings insight into improving future pathological image analysis.

## Acknowledgment

This work was partially supported by the National Natural Science Foundation of China (No. 62271023), the Beijing Natural Science Foundation (No. 7202102), and the Fundamental Research Funds for Central Universities.

## Supplementary Materials

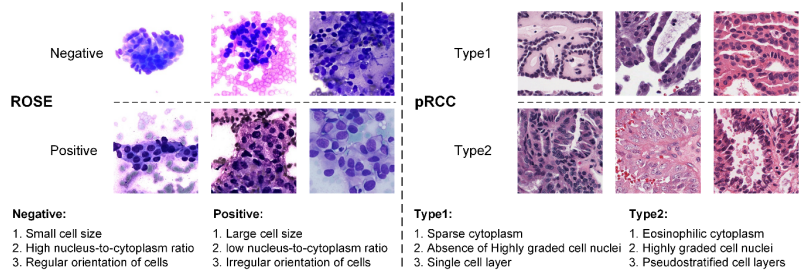

**Fig. 1** Examples of two pathological datasets with their classification criteria.

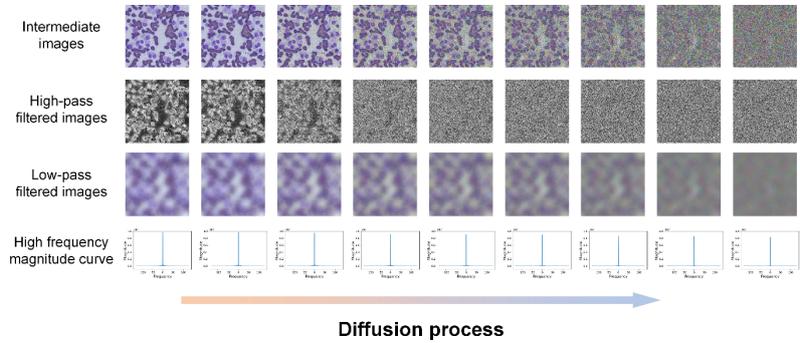

**Fig. 2** Visualization of the diffusion process with FFT.

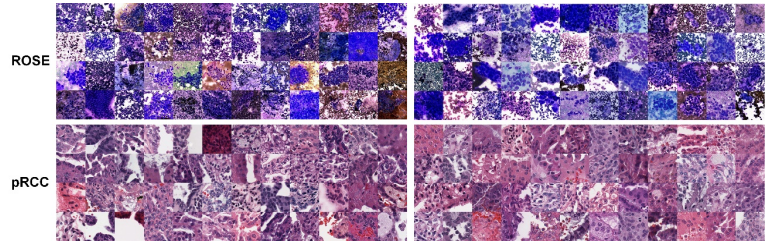

**Fig. 3** Generated samples from IDDPM.

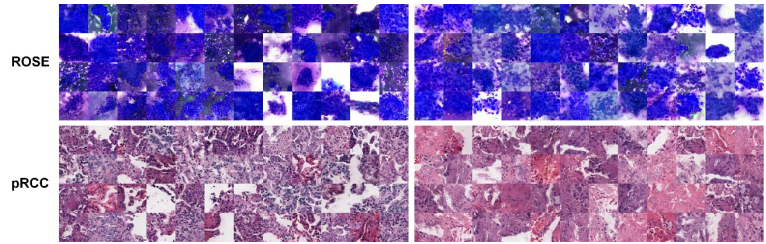

**Fig. 4** Generated samples from LoFGAN.

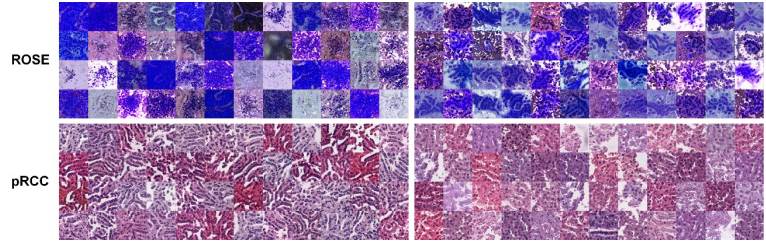

**Fig. 5** Generated samples from MixDL.

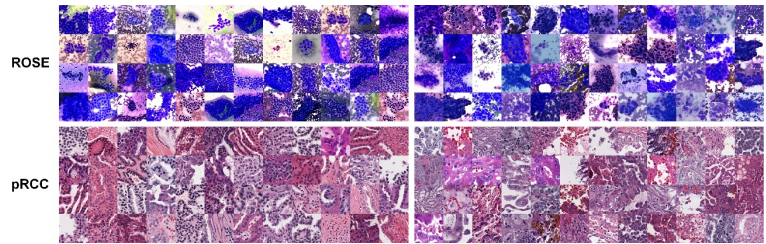

**Fig. 6** Generated samples from ADD (ours).

**Table 1.** Comparisons in the downstream classifications using generated training set. Bold marks the highest of all results, while underlined marks the highest in generated models

| Model | ROSE | | pRCC | |
|---|---|---|---|---|
| | Accuracy (%) | F1-score (%) | Accuracy (%) | F1-score (%) |
| Baseline | **78.13** | **79.09** | 83.50 | 85.36 |
| ProGAN | 64.64 | 67.31 | 71.78 | 72.45 |
| IDDPM | 76.51 | 76.60 | 82.25 | 83.24 |
| LoFGAN | 74.05 | 73.13 | 81.06 | 81.14 |
| MixDL | 61.25 | 60.14 | 68.12 | 70.93 |
| ADD (Ours) | <u>77.92</u> | <u>78.15</u> | **<u>84.04</u>** | **<u>85.49</u>** |

**Table 2.** Comparisons in the downstream classifications using aug-trainset. Bold marks the highest of all results, while the arrow indicates the augmentation of generated models.

| Model | ROSE | | pRCC | |
|---|---|---|---|---|
| | Accuracy (%) | F1-score (%) | Accuracy (%) | F1-score (%) |
| Baseline | 78.13 | 79.09 | 83.50 | 85.36 |
| ProGAN | 74.70 | 73.02 | 75.02 | 75.25 |
| IDDPM | 79.32↑ | 80.81↑ | 83.91↑ | 84.89 |
| LoFGAN | 75.49 | 74.26 | 83.97↑ | 85.55↑ |
| MixDL | 69.66 | 71.83 | 76.28 | 77.26 |
| ADD (Ours) | **80.24↑** | **80.97↑** | **84.76↑** | **86.43↑** |